\documentclass[nouppercase]{kaia}
\usepackage{bm}
\usepackage{amsmath}
\usepackage{bbm}
\usepackage{kotex}
\usepackage{amsthm}
\usepackage{amssymb}
\usepackage{booktabs}
\usepackage{algpseudocode}
\usepackage{algorithm}
\usepackage{comment}

\DeclareMathOperator*{\argmax}{argmax}

\title{Improving Performance of Semi-Supervised Learning by Adversarial Attacks}

\author{Dongyoon Yang}{a}
\author{Kunwoong Kim}{a}
\author{$\text{Yongdai Kim}^{*}$}{a}
\affiliation{Statistics, Seoul National University, Seoul, Korea, \{ydy0415, kwkim.online, ydkim0903\}@gmail.com}{a}

\begin{document}

\maketitle
\begin{abstract}
    Semi-supervised learning (SSL) algorithm is a setup built upon a realistic assumption that access to a large amount of labeled data is tough.
    In this study, we present a generalized framework, named SCAR, standing for Selecting Clean samples with Adversarial Robustness, for improving the performance of recent SSL algorithms.
    By adversarially attacking pre-trained models with semi-supervision, our framework shows substantial advances in classifying images.
    We introduce how adversarial attacks successfully select high-confident unlabeled data to be labeled with current predictions.
    On CIFAR10, three recent SSL algorithms with SCAR result in significantly improved image classification.
    
\end{abstract}
\begin{keywords}
	Semi-supervised Learning, Adversarial Attacks, Image Classification
\end{keywords}

\section{Introduction}

In supervised learning, labeling has emerged as a crucial problem due to its cost, time, and expertise knowledge \cite{Zhu2009IntroductionTS}.
To resolve this issue, various Semi-Supervised Learning (SSL) algorithms, algorithms for learning models only with a small amount of labeled data, are developed and showed successful performances \cite{https://doi.org/10.48550/arxiv.1905.02249, NEURIPS2020_44feb009, https://doi.org/10.48550/arxiv.2001.07685, https://doi.org/10.48550/arxiv.1610.02242, https://doi.org/10.48550/arxiv.1703.01780, https://doi.org/10.48550/arxiv.1704.03976, berthelot2019remixmatch, zhang2021flexmatch}.
To be more rigorous and precise, SSL is built under the assumption that we can only access a few labels.
That is, the number of unlabeled data dominates the number of labeled ones.
In this paper, we propose a strong but simple technique for improving image classification performance via adversarial attacks to a given pre-trained model by semi-supervision.

We begin with revisiting general assumptions and recent algorithms for SSL. Semi-supervised learning algorithms can be constructed successfully under three main assumptions used in general: the \textit{cluster assumption}, the \textit{low-density separation} assumption, and the \textit{manifold assumption} \cite{10.7551/mitpress/9780262033589.001.0001, https://doi.org/10.48550/arxiv.2103.00550}.
The cluster assumption is that two data in the same cluster should belong to the same class or have similar predictions induced by a given classifier.
The low-density separation requires that the decision boundary of the true optimal classifier should lie in a low-density region.
The manifold assumption requires that two similar data on a low-dimensional manifold should be predicted similarly or classified in the same class.
In some sense, it is similar to the cluster assumption but more specific to high-dimensional data on a manifold such as images.

In this study, we introduce SCAR, which stands for selecting clean samples with adversarial robustness, to improve image classification with pre-trained models by semi-supervision.
Particularly, we utilize three SSL algorithms, VAT \cite{https://doi.org/10.48550/arxiv.1704.03976}, MixMatch \cite{https://doi.org/10.48550/arxiv.1905.02249}, and FixMatch \cite{https://doi.org/10.48550/arxiv.2001.07685} to pre-train classifiers, and apply adversarial attacks to select high-confident unlabeled data.
In the following section, we briefly revisit recent SSL algorithms and also introduce adversarial attacks for implementing SCAR.


\section{Preliminaries}

\subsection{Related works: semi-supervised Learning Algorithms}

We revisit popular algorithms and shortly describe the core motivations and philosophies.

Consistency loss-based methods \cite{https://doi.org/10.48550/arxiv.1610.02242, https://doi.org/10.48550/arxiv.1703.01780, https://doi.org/10.48550/arxiv.1704.03976, NEURIPS2020_44feb009} with the manifold assumption adds a consistency loss term in that realistic augmentation of an image should be predicted equal to the original image.
$\Pi-$Model \cite{https://doi.org/10.48550/arxiv.1610.02242} applies random augmentation on an image and minimizes the differences among predictions of the original image and the augmented view.
MeanTeacher \cite{https://doi.org/10.48550/arxiv.1703.01780} is novel in the sense that it applies EMA technique for obtaining a better model.
Virtual Adversarial Training (VAT) \cite{https://doi.org/10.48550/arxiv.1704.03976} attacks adversarially on given images, not the model, and regularizes using the prediction of the adversarial input image.
Unsupervised Domain Adaptation (UDA) \cite{NEURIPS2020_44feb009} extends to text classification using a specific augmentation techniques.

The following four methods combine consistency loss-based and another technique, which results in a great improvement in the performance.
MixMatch \cite{https://doi.org/10.48550/arxiv.1905.02249} uses the entropy minimization with pseudo-labels on unlabeled data obtained by the current learned model.
ReMixMatch \cite{berthelot2019remixmatch} applies two more specific approaches (distribution aligning and augmentation anchoring) to MixMatch.
FixMatch \cite{https://doi.org/10.48550/arxiv.2001.07685} considers high-confident unlabeled data only to avoid degradation of performances due to mislabeled data.
FlexMatch \cite{zhang2021flexmatch} is a diverse of FixMatch with flexible thresholds for selecting high-confident data.

From now on, we further describe the three baseline algorithms used in this paper.

\paragraph{VAT \cite{https://doi.org/10.48550/arxiv.1704.03976}}

Virtual Adversarial Training (VAT) proposes to use adversarial inputs in consistency loss.
That is, it finds an adversarial sample of each input so that the model's predictions should have similar values.
It results in consistency over a neighborhood of each data, which means that all data in the neighborhood should be classified similarly to the original data.

\paragraph{MixMatch \cite{https://doi.org/10.48550/arxiv.1905.02249}}
MixMatch is a novel semi-supervised learning framework in that it proposes to use MixUp \cite{zhang2018mixup} for annotating pseudo-labels to unlabeled data with the two common techniques: entropy minimization and consistency loss.



\paragraph{FixMatch \cite{https://doi.org/10.48550/arxiv.2001.07685}}

FixMatch is a combination of MixMatch and consistency loss-based methods with the threshold for improvement of confidence in pseudo-labels.
The key idea of FixMatch is the following.
For a given data $\bm{x},$ a weak augmentation transformation $t_{weak},$ and a strong augmentation transformation $t_{strong},$ this method annotates a pseudo-label generated by the prediction of $t_{weak}(\bm{x})$ to strongly-augmented unlabeled data $t_{strong}(\bm{x}).$
Here, we only consider high-confident predictions to generate pseudo-labels, i.e., the prediction score of $t_{weak}(\bm{x})$ should be higher than a given threshold, e.g., 0.95.

\subsection{Notations}
Let $\mathcal{X} \subset \mathbb{R}^d$ be the input space, $\mathcal{Y} = \left\{1, \cdots, C\right\}$
be the set of output labels and $f_{\bm{\theta}} : \mathcal{X} \rightarrow \mathbb{R}^{C}$ be the score function parametrized by neural network parameters $\bm{\theta}$ such that $\mathbf{p}_{\theta}(\cdot|\bm{x}) =\operatorname{softmax}(f_{\bm{\theta}}(\bm{x})) \in \mathbb{R}^C$ is the vector of the conditional
class probabilities. Let $F_{\bm{\theta}}(\bm{x}) = \underset{c}{\argmax} f^{c}_{\bm{\theta}}(\bm{x}),$  
$\mathcal{B}_{p}(\bm{x}, \varepsilon) = \left\{\bm{x}' \in \mathcal{X} : \lVert \bm{x}- \bm{x}' \rVert_p \leq \varepsilon \right\}$ and $\mathbbm{1}(\cdot)$ be the indicator function.

We write the labeled dataset as $\mathcal{D}_{l}=\left\{(\bm{x}_i, y_i) \in \mathbb{R}^{d+1}: i=1, \cdots, n_{l} \right\}$, and
unlabeled dataset as $\mathcal{D}_{ul}=\left\{\bm{x}_j \in \mathbb{R}^{d}: j=1, \cdots, n_{ul} \right\}.$

For the three SSL methods, the overall loss function is in common formulated as
\begin{equation}\label{loss}
    \begin{split}
        \mathcal{L}(\theta) = \mathcal{L}_{labeled}(\theta) + \lambda \mathcal{L}_{unlabeled}(\theta)
    \end{split}
\end{equation}
for some hyperparameter $\lambda > 0,$
where $\mathcal{L}_{labeled}(\theta)$ is the supervised loss with labeled data and
$\mathcal{L}_{unlabeled}(\theta)$ is the regularization loss (for consistency or/and entropy minimization) with unlabeled data.

We denote those algorithms as $\ell_{\text{semi}}(\mathcal{D}_{l}, \mathcal{D}_{ul}),$ which solves the loss function in equation (\ref{loss}) with given iterations (i.e., epochs).


\subsection{Adversarial Attacks}
The adversarial example with maximum perturbation $\varepsilon$ is defined by 
\begin{equation}
\label{adv-example}
\bm{x}^{\text{adv}}= \underset{\bm{x}' \in \mathcal{B}_p(\mathbf{X}, \varepsilon) }{\argmax\;\;\;} \mathbbm{1}\left\{F_{\theta}(\mathbf{X'}) \neq \mathbf{Y} \right\}.
\end{equation}
The most popular method for finding $\bm{x}^{\text{adv}}$ is PGD \cite{madry2018towards}.
We can get adversarial examples when $p = \infty$ as follows :
\begin{equation}
\label{pgd}
    \bm{x}^{(t+1)}=\bm{\Pi}_{\mathcal{B}_{p}(\bm{x}, \varepsilon) }\left(\bm{x}^{(t)} + \alpha \operatorname{sgn}\left(\nabla_{\bm{x}^{(t)}} \ell(f_{\bm{\theta}}(\bm{x}^{(t)}),y)\right)\right)
\end{equation}
where $t= 1, \cdots, T$ and $\bm{\Pi}_{\mathcal{B}_{p}(\bm{x}, \varepsilon)}(\cdot)$ is the projection operator to $\mathcal{B}_{p}(\bm{x}, \varepsilon)$ and $\bm{x}^{(0)}=\bm{x}$.
\section{Proposed Method}
We propose a Selecting Clean samples with Adversarial Robustness (SCAR) algorithm, which selects clean samples in unlabeled data by adversarial attacks and labels with predictive classes.

\subsection{Motivation}
In \cite{https://doi.org/10.48550/arxiv.2001.07685}, the confidence $\underset{y \in \mathcal{Y}}{\max} p_{\bm{\theta}}(y|\bm{x}_i)$ is used for selecting unlabeled samples to be labeled with pseudo label. If $\underset{y \in \mathcal{Y}}{\max} p_{\bm{\theta}}(y|\bm{x}_i) > \tau $, that unlabeled data is labeled to be  $\underset{y \in \mathcal{Y}}{\argmax}\; p_{\bm{\theta}}(y|\bm{x}_i)$.
Motivated from this perspective, we can replace the criterion $\mathbbm{1} \{ \underset{y \in \mathcal{Y}}{\max} p_{\bm{\theta}}(y|\bm{x}_i) > \tau \}$ in \cite{https://doi.org/10.48550/arxiv.2001.07685} with
\begin{equation}
\small
\label{eqn3}
\mathbbm{1} \{ \underset{y \in \mathcal{Y}}{\argmax} p_{\bm{\theta}}(y|\bm{x}_i)  = \underset{y \in \mathcal{Y}}{\argmax} \underset{\bm{x}'_i \in \mathcal{B}_{p}(\bm{x}_i, \varepsilon)}{\min} p_{\bm{\theta}}( \underset{y \in \mathcal{Y}}{\argmax} p_{\dot{\bm{\theta}}}(y|\bm{x}_i) |\bm{x}'_i)  \},
\end{equation}
where $\dot{\bm{\theta}}$ is a copy of $\bm{\theta}$, but it is fixed when finding $\bm{x}'_i$. To find $\bm{x}'_i$ satisfying (\ref{eqn3}), we use the adversarial attack. In summary, the unlabeled data with adversarial robustness are labeled to be predictive class.

\subsection{Selecting Clean samples with Adversarial Robustness (SCAR)}
The procedure of SCAR is summarized in lines 5 - 9 of Algorithm \ref{scar:alg}. First, we define the adversarial attack with perturbation size $\varepsilon$ in (\ref{adv-example}). Second, make pseudo labels for unlabeled data by pre-trained model parametrized $\tilde{\bm{\theta}}$. Lastly, the data with adversarial robustness (the samples with robust to adversarial attack) is considered to explore clean samples and add them to labeled data set. In this procedure, sensitivity \ref{sensitivity} and specificity \ref{specificity} has trade-off of $\varepsilon$. For large $\varepsilon$, the sensitivity is high, and the specificity is low since the stronger attack pushes decision boundary of samples. We can observe the trade-off in Table \ref{table2}.

\begin{algorithm}[t!]
    \caption{SCAR-Algorithm}
    \label{scar:alg}
    \textbf{Input} : network $f_{\bm{\theta}}$,\\
    labeled dataset $\mathcal{D}_{l}=\left\{(\bm{x}_i, y_i) \in \mathbb{R}^{d+1}: i=1, \cdots, n_{l} \right\}$,\\ 
    unlabeled dataset $\mathcal{D}_{ul}=\left\{\bm{x}_j \in \mathbb{R}^{d}: j=1, \cdots, n_{ul} \right\}$,\\
    semi-supervised loss function $\ell_{\text{semi}}(\mathcal{D}_{l}, \mathcal{D}_{ul}; f_{\theta})$,\\
    FGSM with maximum perturbation size $\epsilon$ $\mathcal{A}_{\varepsilon}(\bm{x}_i, y_i, f_{\bm{\theta}})$,\\
    learning rate $\eta$,  number of epochs $T$, number of batch $B$, batch size $K$. \\
    \textbf{Output} : network $f_{\bm{\theta}}$
    \begin{algorithmic}[1]
    \State Pre-train $f_{\tilde{\bm{\theta}}}$ by semi-supervised Learning Algorithm with $\ell_{\text{semi}}(\mathcal{D}_{l}, \mathcal{D}_{ul})$.
    \State Set $\bm{\theta}_{0} = \tilde{\bm{\theta}}$
    \For{$ t = 1 , \cdots, T$}
        \For{$ j = 1 , \cdots, n_{ul}$}
            \State $\tilde{y}_j = F_{\tilde{{\bm{\theta}}}}(\bm{x}_{j})$
            \State $\bm{x}^{\text{adv}}_j = \mathcal{A}_{\varepsilon}(\bm{x}_j, \tilde{y}_j; f_{{\tilde{\bm{\theta}}}})$
            \State $\tilde{y}_j^{\text{adv}} = F_{{\tilde{\bm{\theta}}}} (\bm{x}^{\text{adv}}_j) $
            \If{$\tilde{y}_j = \tilde{y}_j^{\text{adv}}$}
                \State $\mathcal{D}_{l,t} = \mathcal{D}_l \cup \{\bm{x}_j, \tilde{y}_j \}$
            \EndIf
            \State $\bm{\theta}_{t} = \bm{\theta}_{t-1} - \eta \ell_{\text{semi}}(\mathcal{D}_{l,t}, \mathcal{D}_{ul}; f_{\bm{\theta}_t} )$
        \EndFor
    \EndFor
    \end{algorithmic}
    \textbf{Return} $f_{\bm{\theta}}$
\end{algorithm}

\section{Experiments}


\begin{table}[t!]
    \centering
    \begin{tabular}{c|c}
    \toprule
    \textbf{Method} & \textbf{Acc.}  \\
    \midrule
    \midrule 
    VAT \cite{https://doi.org/10.48550/arxiv.1704.03976} &  85.25 \\
    VAT + SCAR      &       \textbf{86.79}  \\
    \midrule
    MixMatch \cite{https://doi.org/10.48550/arxiv.1905.02249}    & 89.99 \\
    MixMatch + SCAR  & \textbf{92.99} \\
    \midrule
    FixMatch \cite{https://doi.org/10.48550/arxiv.2001.07685}  & 94.13 \\
    FixMatch + SCAR  &  \textbf{95.17} \\
    \bottomrule
    \end{tabular}
    \caption{\textbf{Performance on CIFAR10.}}
    \label{table1}
\end{table}

\subsection{Dataset}

We use CIFAR10 \cite{krizhevsky2009learning}, which consists of 60,000 images with 32 by 32 size in 10 classes.
60,000 images are split into 50,000 training and 10,000 test image samples, and we only use randomly selected 4,000 images for labeled data from 50,000 training images. 
For the pre-processing, the images are normalized into [0, 1].

\subsection{Performance}
In Table \ref{table1}, we compare the accuracy of semi-supervised learning with and without SCAR. 
We pre-train a network with three existing SSL algorithms with fixed epochs, select the last model, and report the accuracies of the last model.
Using the selected model, we apply SCAR for the same epochs more and report the final accuracies computed on the final model selected during SCAR training procedure.
For the epochs, we set  $200$ and report the results in Table \ref{table1}.
We set $\lambda = 1.0, $ and $\lambda = 0.75$ for FixMatch and MixMatch, respectively.

\subsection{Detecting the correctly labeled data}
We implement an ablation study to know how our method correctly picks out the clean label samples. 
The sensitivity and specificity are defined as

\begin{equation}
\label{sensitivity}
    \text{Sensitivity} := \dfrac{\lvert \mathbbm{1}\left\{y_i=F_{\bm{\theta}}{(\bm{x}_i)} = F_{\bm{\theta}}{(\bm{x}^{\text{adv}}_i)}   \right\}  \rvert}{\lvert \mathbbm{1}\left\{F_{\bm{\theta}}{(\bm{x}_i)} = F_{\bm{\theta}}{(\bm{x}^{\text{adv}}_i)}   \right\}  \rvert},     
\end{equation}
and
\begin{equation}
\label{specificity}
    \text{Specificity} := \dfrac{\lvert \mathbbm{1}\left\{y_i \neq F_{\bm{\theta}}{(\bm{x}_i)}  \right\} \mathbbm{1}\left\{F_{\bm{\theta}}{(\bm{x}_i)} \neq F_{\bm{\theta}}{(\bm{x}^{\text{adv}}_i)}   \right\}  \rvert}{\lvert \mathbbm{1}\left\{F_{\bm{\theta}}{(\bm{x}_i)} \neq F_{\bm{\theta}}{(\bm{x}^{\text{adv}}_i)}   \right\}  \rvert}.
\end{equation}
The sensitivity and specificity mean the rates of success of prediction about correctly and incorrectly pseudo-labeling, respectively.


In Table \ref{table2}, VAT ($\varepsilon=1$)
shows insufficient performance in terms of sensitivity compared to MixMatch ($\varepsilon=1$) or
FixMatch ($\varepsilon=1$). For achieving similar sensitivity to other methods, a large $\varepsilon$ is necessary. But, in this case, the number of unlabeled data to be selected is decreasing. This phenomenon is quite reasonable since adversarial robustness is utilized for regularization in VAT. Also, we can see that selection of optimal $\varepsilon$ in every method is crucial for better performance.
\begin{table}[h]
\small
\label{table2}
\begin{center}
\begin{tabular}{|l|c|c|}
\hline
Learning Method & Sensitivity (\%) & Specificity (\%)\\
\hline\hline
VAT ($\varepsilon=1$) & 92.22 (7587/8227) & 47.10 (835/1773) \\
VAT ($\varepsilon=2$) & 95.76 (6543/6833) & 37.42 (1185/3167) \\
VAT ($\varepsilon=3$) & 97.48 (5494/5636) & 30.55 (1333/4364) \\
VAT ($\varepsilon=4$) & 98.35 (4716/4795) & 26.81 (1396/5205) \\
VAT ($\varepsilon=6$) & 98.93 (3684/3724) & 22.86 (1435/6276) \\
\hline
MixMatch ($\varepsilon=1$) & 98.91 (6054/6124) & 24.13 (936/3879) \\
FixMatch ($\varepsilon=1$)  & 99.59 (6385/6411) & 15.63 (561/3589) \\
\hline
\end{tabular}
\end{center}
\caption{\textbf{Sensitivity and Specificity}. We train models for each learning method and present the sensitivity \ref{sensitivity} and specificity \ref{specificity}. for test set. The numbers in the brackets are counting numbers for each measurement.}
\label{table2}
\end{table}

\section{Conclusion}

In this paper, we propose a generalized fine-tuning framework called SCAR, which enhances the performance of trained networks by semi-supervision in image classification.
Among various algorithms for SSL, VAT, MixMatch, and FixMatch are used to pre-train classifier networks.
We experimentally analyze that those SSL approaches with SCAR would improve the performances with significant margins.
SCAR can be interpreted as a fine-tuning framework for selecting high-confident unlabeled data, which induces a two-step scheme for obtaining the final trained model.

We also note three topics for future work here.
First, instead of the two-step procedure of SCAR, it would receive more interest if one could construct an end-to-end learning procedure of SCAR, that is, providing high-confident labeled data using adversarial attacks during training time.
Second, our work is based on only CIFAR10.
Thus, it would enhance the novelty of SCAR that the consistent results on various image datasets such as SVHN, CIFAR100, and STL10. 
Lastly, the SCAR can be applied to more semi-supervised learning algorithms such as UDA \cite{NEURIPS2020_44feb009}, ReMixMatch \cite{berthelot2019remixmatch} and FlexMatch \cite{zhang2021flexmatch}, which is state-of-the-art. If SCAR is working for the most of semi-supervised learning algorithms, it boosts the novelty.

\end{document}